# Defense against adversarial attacks on deep convolutional neural networks through nonlocal denoising

**Sandhya Aneja[1], Nagender Aneja[2], Pg Emeroylariffion Abas[1], Abdul Ghani Naim[2]**
[1]Faculty of Integrated Technologies, Universiti Brunei Darussalam, Bandar Seri Begawan, Brunei Darussalam
[2]School of Digital Science, Universiti Brunei Darussalam, Bandar Seri Begawan, Brunei Darussalam

**ABSTRACT**

Despite substantial advances in network architecture performance, the susceptibility of adversarial attacks makes deep learning challenging to implement in safety-critical applications. This paper proposes a data-centric approach to addressing this problem. A nonlocal denoising method with different luminance values has been used to generate adversarial examples from the Modified National Institute of Standards and Technology database (MNIST) and Canadian Institute for Advanced Research (CIFAR-10) data sets. Under perturbation, the method provided absolute accuracy improvements of up to 9.3% in the MNIST data set and 13% in the CIFAR-10 data set. Training using transformed images with higher luminance values increases the robustness of the classifier. We have shown that transfer learning is disadvantageous for adversarial machine learning. The results indicate that simple adversarial examples can improve resilience and make deep learning easier to apply in various applications.

*Corresponding Author:*

Nagender Aneja
School of Digital Science, Universiti Brunei Darussalam
Bandar Seri Begawan, Brunei Darussalam
Email: nagender.aneja@ubd.edu.bn

## 1. INTRODUCTION

Machine learning helps address the challenges of different industries, including transportation [1], [2], cybersecurity [3]–[7], retail [8], smart home [9], social networks [10], health sciences [11], [12], fake news detection [13], and financial services sectors [14]. In particular, deep learning is advantageous in classification problems involving image recognition [15], object recognition, speech recognition, and language translation [16] to give better classification performance. However, recent research has indicated that deep learning algorithms are prone to attacks and can be manipulated to influence algorithmic output [17]–[20]. Evtimov *et al*. [21] demonstrated an adversarial attack to manipulate an autonomous vehicle or a self-driving car to misclassify stop signs such as the speed limit, a significant concern in the rapidly developing autonomous transportation domain. In cybersecurity, Kuchipudi *et al.* [22] and Yang *et al*. [23] have shown that deep learning based spam filters and artificial intelligence based malware detection tools, respectively, can be bypassed by deploying adversarial instruction learning approaches.

Similarly, despite advanced progress in face recognition and speech recognition that led to their deployment in real-world applications, such as retail, social networks, and intelligent homes, Vakhshiteh *et al*. [24] and Schonherr *et al*. [25] have demonstrated their vulnerabilities against numerous attacks to illustrate potential research directions in different areas. Even in the area of finance and health, which traditionally requires a high level of robustness in the system, adversarial attacks are capable of manipulating the system, for example, by deceiving fraud detection engines to register fraudulent transactions [26], manipulating the health status of individuals [27], and fooling text classifiers [28]–[31].

Algorithmically crafted perturbations, no matter how small, can be used as adversarial instructions to manipulate the classification results of deep neural network (DNN)-based image classifiers, with some of these adversaries capable of finding the relevant perturbations without access to the network architecture. Goodfellow *et al*. [18] demonstrate a picture of a panda classified as a gibbon, with the addition of perturbations in the picture, despite human eyes still perceiving the picture as that of a panda. Similarly, Evtimov *et al*. [21] have demonstrated that subtle perturbations, unnoticed by the naked eye, can alter the result of the DNN-based image classifier from that of a stop sign to a speed limit. It can be argued that these attacks might be challenging to execute on a real-time system, such as an autonomous vehicle since they require the real-time images obtained from the vehicle sensors to be intercepted and then fed to the perturbations before being passed to the classifier. However, the presence of large types of attacks proposed by several researchers requires the development of robust attack-agnostic defense mechanisms against these attacks [32]–[34].

Figure 1 illustrates the fast gradient sign attack [18] in the Modified National Institute of Standards and Technology database (MNIST) data set, with the eps value reflecting the proportion of perturbation of the fast gradient sign method added to the image. Without perturbation, that is, eps: 0, all numbers are correctly classified, whereby the notation 3 → 3 indicates that the number 3 has been correctly classified as the number 3. However, as the perturbation is added, some numbers are incorrectly classified. In Figure 1, the numbers 2, 6, 2, 9, and 2 have been incorrectly classified as 8, 8, 8, 4, and 7. Goodfellow *et al*. [18] have shown that the accuracy of classification models decreased as the perturbation value increased. Hendrycks *et al*. [35] curated two data sets, ImageNet-A and ImageNet -O, to function as natural adversarial example test sets. It has been demonstrated that the performances of different machine learning models deteriorate significantly from adversarial attacks, with the DenseNet-121 model showing 2% and near-random chance level accuracies on the adversarial data sets ImageNet-A and ImageNet-O, respectively. Thus, a robust mechanism is required to protect against these attacks.

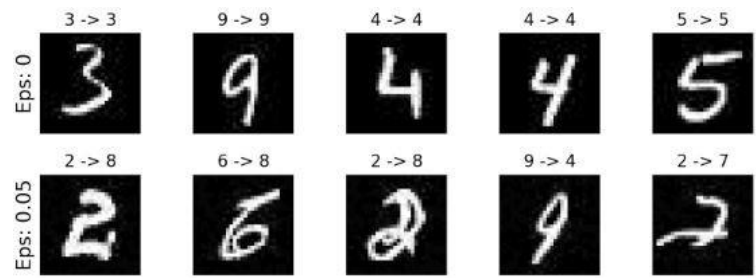

Figure 1. Fast Gradient Sign Attack on MNIST

Generally, two approaches can be adopted to improve the robustness of a classification model against adversarial attacks: the data-centric approach and the model-centric approach [36]. In a data-centric approach, data are the primary asset to increase the robustness of a classification model while keeping the algorithm, architecture, and hyperparameters constant. The data-centric approach requires the creation of an additional data set to improve the performance of a given fixed model. On the other hand, the model and its parameters vary in a model-centric approach, with the training data constant. Yan *et al*. [37] illustrated a model-centric approach by integrating an adversarial perturbation-based regularizer into the classification objective. The authors have demonstrated that the addition of the regularizer can significantly outperform other methods in terms of robustness and accuracy. Using multi layer perceptron (MLP) on the MNIST data set, the resultant model provides more than two times better robustness to DeepFool adversarial attack, and an improvement in accuracy of 46.69% during fast gradient signed (FGS) attack, over traditional MLP; illustrating the capability of DNN model with the regularizer in learning to resist potential attacks, directly and precisely.

Similarly, Amini *et al*. [38] presented evidential models that capture the increased uncertainty in samples that have been adversarially perturbed. Subsequently, the evidential deep learning method learned a grounded representation of the uncertainty of the data without the need for sampling. Specifically addressing gradient-based adversarial attacks, Carbone *et al*. [39] proved that Bayesian neural networks (BNNs) trained with Hamiltonian Monte Carlo (HMC) and variational inference could provide a robust and accurate solution to such attacks in a suitably defined large data limit.

Xie *et al*. [40] presented the network architecture, which uses denoise image features using non-local means to improve robustness. The authors have demonstrated a 55.7% accuracy on the ImageNet data set under white-box attacks, a substantial improvement in accuracy over the 27.9% accuracy using the traditional method. The authors proposed a network architecture with blocks to denoise the features using non-local means. The denoise operation of the authors is processed by 1×1 convolutional layer and then added to the input of the block by a residual connection. Xie *et al*. [41] proposed using adversarial examples in the training data set with different auxiliary batch norms to improve ImageNet trained models. The auxiliary batch norm has marked differences compared to a normal example. The EfficientNet-B7 architecture has been used in ImageNet, with an improvement of 0.7% accuracy shown to increase robustness under different attacks.

Moosavi-Dezfooli *et al*. [42] proposed dividing the input image into multiple overlapping patches, further denoising each patch independently and reconstructing the denoised image by averaging the pixels in overlapping patches. The authors set the overlap to 75% of the patch size. Liao *et al*. [43] proposed a high-level representation-guided denoiser (HGD) defense for image classification. The authors used the loss function as the difference between the top-level outputs of the target model induced by the original and adversarial examples instead of denoising pixels. The authors used a UNet-like autoencoder for denoising. Chow *et al*. [44] proposed to combine the unsupervised model denoising ensemble with the supervised model verification ensemble. The authors used denoising ensembles to remove different types of noise and create the number of denoised images. Noise is removed with the help of an auto-encoder. The verification ensemble then votes on all denoised images. Thang and Matsui [45] used image transformation and filter techniques to identify adversarial examples sensitive to geometry and frequency and to remove adversarial noise.

Most of the prior publications have considered denoising a promising approach for adversarial robustness. However, previous published studies have considered a model-centric approach in which the emphasis is on developing different architectures or tuning various hyperparameters. However, this paper studies the performance of a classification model against attack by adding adversarial examples in the training data set and keeping the architecture and hyperparameters constant, hence falling under the data-centric approach.

## 2. METHOD

This paper considers a data-centric solution to the MNIST [46] and Canadian Institute for Advanced Research (CIFAR-10) [47] data sets. The MNIST data set consists of 60,000 training images and 10,000 test images, with images comprised of handwritten grayscale digits between 0 and 9 and of size 28×28 pixels. The images in the data set are centered and normalized so that the pixel value is in the range of [0, 1]. On the other hand, the CIFAR-10 data set consists of 60,000 colored images of 10 classes of items: airplane, automobile, bird, cat, deer, dog, frog, horse, ship, and truck. There are 6,000 images per class, and of the 60,000 colored images, 50,000 are training images and 10,000 are test images. Each image has a size of 32×32 pixels.

In this study, three additional data sets using the nonlocal denoising method were created from the MNIST and CIFAR-10 data sets. The algorithm selects a pixel, takes a small window around the selected pixel, and then performs a scan for similar windows around the original image to average all windows. The average value, which represents estimates of the original image after noise suppression, is then used to replace the pixel value. Given that a grayscale image from the MNIST data set is represented by $I^g$ and a colored image from the CIFAR-10 data set is represented by $I^c$, where $c \in \{R, G, B\}$ represents the red, green, and blue channels, respectively. The gray value of image $I^g$ and the color value of image $I^c$ in pixel p may be represented by $I^g(p), p \in I^g$ and $I^c(p), p \in I^c$ respectively. For simplicity, c can be used to describe either the three color channels of a colored image, that is, $c \in \{R, G, B\}$, or the grayscale image, i.e. c = g, so that $I^c$ can be interchangeably used to represent both grayscale and colored images. The respective estimates $\hat{I}^c(p)$ of the original image $I^c$ at pixel p using the nonlocal denoising method may be represented by $\hat{I}^c(p) = \frac{1}{C(p)} \sum_{q \in \Omega(p)} I^c(p) W(p, q)$, where Ω(p) is a square patch of size 21×21 centered at p, representing the search zone for the nonlocal denoising filter around the vicinity of the pixel p. C(p) represents the normalizing factor of pixel p and W (p, q) denotes the weighting factor of pixel q on pixel p. Both parameters are related by $C(p) = \sum_{q \in \Omega(p)} W(p, q)$.

The weighting factor W(p, q) of the pixel q on p is calculated as $W(p,q) = e^{-\frac{max(d^2(w(p),w(q)) - 2\sigma^2, \ 0)}{h^2}}$, where h represents the luminance or color component, and σ denotes the standard deviation of the noise. The large value of h removes more noise from the image; however, it may also decrease its quality. $d^2(w(p), w(q))$ is the squared Euclidean distance between the two square patches of size 7×7 centered at p and q, i.e. ω(p) and ω(p) which can be calculated from $d^2(w(p), w(q)) =$ $\frac{1}{n(c) \times 7^2} \sum_c \sum_{j \in \Omega(0)} [I^c(p+j) - I^c(q+j)]^2$, where n(c) represents the number of elements in c, i.e. n(c) = 1 for the grayscale image, and n(c)=3 for the colored image. In this paper, three values of h were considered, h ∈ {3, 5, 15}; representing the values of small, medium, and large luminance/color components, respectively, to produce three additional data sets from the MNIST and CIFAR-10 data sets. These additional data sets formed adversarial examples and were used to train the classifiers, in addition to the original training data set. ResNet18 and ResNet50 have been used for the MNIST and CIFAR-10 data sets, respectively. Models pretrained and non-pretrained on ImageNet have been utilized for comparative purposes.

ResNet18 is a residual network with 18 layers, consisting of 17 convolutional layers and one fully connected layer. The residual network is comparatively different from the traditional neural network in which each layer feeds to the next layer. In a residual network, skip connections with double or triple layer skips are used, containing nonlinear (ReLU) activations and batch normalization in between. Thus, the residual network feeds to the next layer and layers 2-3 hops away. The concept of a skip connection is based on brain structure; for example, neurons in cortical layer VI receive input from layer I and thus skip intermediary layers [48]. Skipping connections solve the problem of vanishing gradients and accuracy saturation.

ResNet50, also a residual network, but with 50 layers deep, has been used for the CIFAR-10 data set. The more complex and deeper network used for the classification task of the CIFAR-10 data set reflects the more complex data present in the CIFAR-10 data set. The stochastic gradient descent function has been used as an optimizer with a learning rate of 0.01, a momentum of 0.9, and a weight decay of $5e^{-4}$. CosineAnnealingLR scheduler is also added to update the learning rate based on the difference between the target and actual examples. The training was carried out for 30 epochs with a batch size of 256. The perturbation values varied from [0, 0.05, 0.1, 0.15, 0.2, 0.25, 0.3].

## 3. RESULTS AND DISCUSSION

This section presents experimental results and analysis on the effect of using additional data sets generated using the nonlocal denoising method from the MNIST and CIFAR-10 data sets on the robustness of the classifiers. Different luminance values, i.e., h={3, 5, 15}, have been used to generate the data sets to form adversarial examples for additional training data sets. In particular, the performances of the classifiers trained using the data sets generated with the nonlocal denoising method are compared with the classifier trained with the original training data set. Experimental results of all approaches have been shown for the model pretrained using transfer learning and the non-pretrained model.

Figures 2 and 3 present the performance of the ResNet18 classifier in the MNIST data set for different perturbation values. The ResNet18 classifier in Figure 2 has used ImageNet transfer learning for its initial weights. In contrast, the ResNet18 classifier in Figure 3 has been trained without transfer learning, but using only the architecture with initial weights from a gaussian distribution. In Figure 2, it can be seen that the accuracy decreases with an increase in the perturbation. A drop in accuracy of about 60.6% with a perturbation value of 0.3 is shown for the classifier trained on the original training data set, while the classifier trained with images transformed using the nonlocal denoising method with luminance component 15 shows a 51.3% drop in accuracy; a gain of around 9.3% despite keeping the algorithm, and hyperparameters unchanged. A similar reduction in classification accuracy with increased perturbation can be seen in Figure 3 with the non-pretrained ResNet18 classifier. A drop in accuracy of about 48.9% with a perturbation value of 0.3 is shown for the classifier trained on the original training data set. This contrasts with a drop of 44.7% with similar perturbation values for the classifier trained with adversarial examples with luminance component 3, which represents an improvement of 4.2% over the classifier trained with the original training data set.

Comparison between the pretrained and non-pretrained ResNet18 classifiers using transfer learning indicates that using transfer learning puts the classifier into a worse position for adversarial attacks with a more significant drop in accuracy in the advent of attacks. It is also noted that the images transformed using the nonlocal method with the luminance component 3 perform marginally better (3.2% difference) than with luminance component 15 for the non-pretrained ResNet18 classifier in Figure 3. In contrast, for the pretrained ResNet18 classifier in Figure 3, the higher luminance component performs better (7.6% difference) than the lower luminance component.

Figures 4 and 5 present the performance of the ResNet50 classifier in the CIFAR-10 data set for different perturbation values. The ResNet50 classifier has used transfer learning with initial weights from ImageNet in Figure 4, while the ResNet50 classifier in Figure 5 has used initial weights from a Gaussian distribution only. A general drop in accuracy can be observed with increasing perturbation value in both Figures 4 and 5. For the ResNet50 classifier using transfer learning in Figure 4, it can be seen that the accuracy drops by approximately 73% and 60% with a perturbation value of 0.3 for the classifier trained with the original training data set only and with images transformed using the nonlocal denoising method with luminance component 15, respectively. This illustrates that training using the transformed images improves accuracy performance by 13% while keeping the algorithm and hyperparameters unchanged. Figure 5 shows accuracy drops of around 66% and 56% with a perturbation value of 0.3 for classifiers trained with the original training data set only and with transformed images with luminance component 15, respectively. Again, this illustrates the advantage of using transformed images using the nonlocal denoising method as adversarial examples in the training data set, albeit with a lower improvement of only 4%. Thus, the classifier trained with transformed images using the nonlocal denoising method is generally better than the classifier trained only with the original training data set. This is true for both ResNet50 classifiers utilizing transfer learning, i.e., pretrained and non-pretrained models, in Figures 4 and 5, respectively.

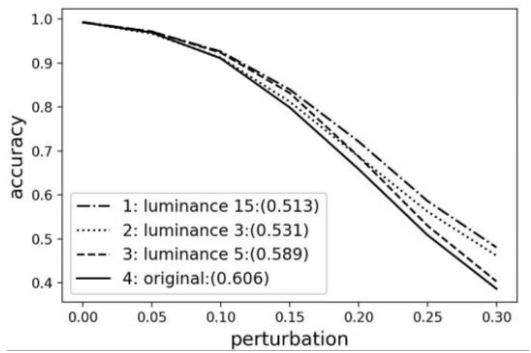
Figure 2. MNIST (pretrained) with ResNet18

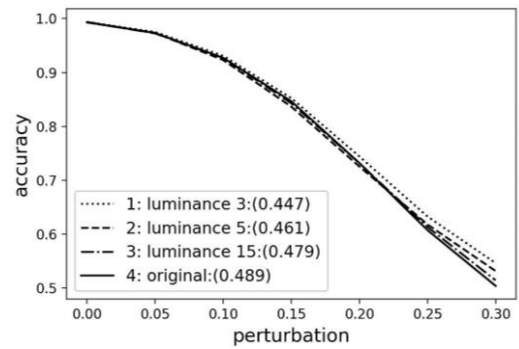
Figure 3. MNIST (not pretrained) with ResNet18

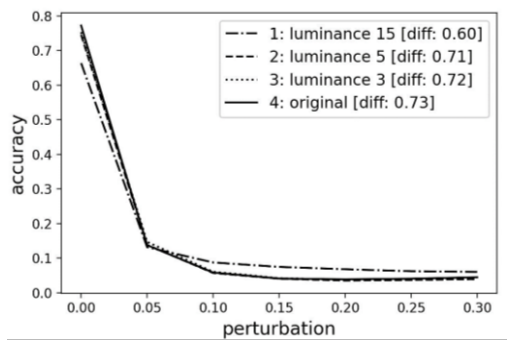
Figure 4. CIFAR-10 (pretrained) with ResNet50

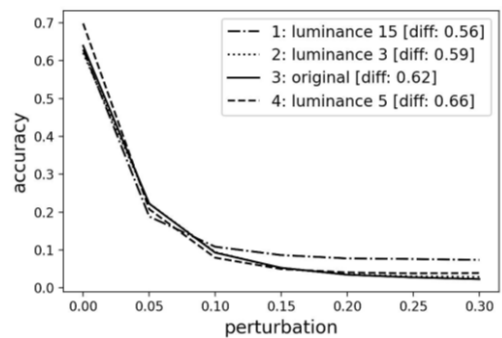
Figure 5. CIFAR-10 (not pretrained) with ResNet50

Comparing Figures 4 and 5 shows that using transfer learning in the ResNet50 classifier provides higher accuracy in the absence of perturbation. However, the perturbation affects the pretrained model more than the non-pretrained model, resulting in a more significant drop of accuracy as more perturbation is introduced. This indicates that the pretrained model which utilizes transfer learning, worsens the robustness of the classifier. Overall, these results indicate that introducing adversarial examples in the training data set, by utilising transformed images using the nonlocal denoising method with high luminance value, without pretraining the network via transfer learning, is advantageous for adversarial machine learning.

## 4. CONCLUSION

Convolutional networks are prone to adversarial attacks, which present a challenge to safety-critical domains where calibrated, robust, and efficient measures of data uncertainty are crucial. Several different deep learning techniques are summarized in this paper to improve the robustness of a classification model even in the presence of adverse perturbations. Two general approaches may be adopted for this purpose: data- centric and model-centric approaches. In this paper, a data-centric approach has been demonstrated, in which the MNIST and CIFAR-10 data sets have been craftily perturbed using the nonlocal denoising method with different luminance values. Experimental results have indicated that the introduction of transformed images as adversarial examples in the training data set is capable of increasing the robustness of the classification model. The method has been shown to provide absolute accuracy improvements of up to 9.3% and 13% on the MNIST and CIFAR-10 data sets, respectively, over classifier trained on the original data sets only, under perturbations. Introducing transformed images with high luminance values gives a more robust classifier. Furthermore, it has been demonstrated that the use of transfer learning is disadvantageous for adversarial machine learning. Future work may include generative adversarial examples for ImageNet and generating denoised adversarial examples using generative adversarial networks. Future work may also involve exploring the effectiveness of Bayesian inference for adversarial defense.

## ACKNOWLEDGEMENT

The authors acknowledge the support given by the SDS Research Grant of Universiti Brunei Darussalam awarded to Dr. Nagender Aneja and Dr. Sandhya Aneja, grant number UBD/RSCH/1.18/FICBF(b)/2021/001.

## BIOGRAPHIES OF AUTHORS

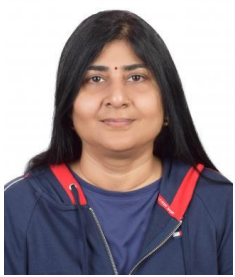

**Dr Sandhya Aneja** is working as Assistant Professor of Information and Communication System Engineering at the Faculty of Integrated Technologies, Universiti Brunei Darussalam. Her primary areas of research interest include wireless networks, high-performance computing, internet of things, artificial intelligence technologies-machine learning, machine translation, deep learning, data science, and data analytics. Further info on her website: https://sandhyaaneja.github.io She can be contacted at email: sandhya.aneja@gmail.com.

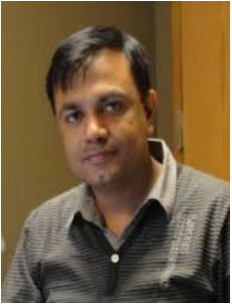 **Dr Nagender Aneja** is working as Assistant Professor at School of Digital Science, Universiti Brunei Darussalam. He did his Ph.D. in Computer Engineering from J.C. Bose University of Science and Technology YMCA, and M.E. Computer Technology and Applications from Delhi College of Engineering. He is currently working in deep learning, computer vision, and natural language processing. He is also the founder of ResearchID.co. Further info on his website: http://naneja.github.io. He can be contact at email: naneja@gmail.com.

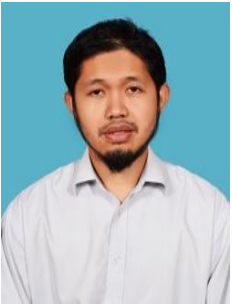 **Pg Dr Emeroylariffion Abas** received his B.Eng. Information Systems Engineering from Imperial College, London in 2001, before obtaining his Ph.D. Communication Systems in 2005 from the same institution. He is now working as an Assistant Professor in General Engineering, Faculty of Integrated Technologies, Universiti Brunei Darussalam. His present research interest are data analytic, energy systems and photonics. Further info on his homepage: https://expert.ubd.edu.bn/emeroylariffion.abas. He can be contacted at email: emeroylariffion.abas@ubd.edu.bn.

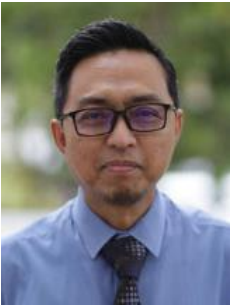 **Dr Abdul Ghani Naim** is a Senior Assistant Professor at School of Digital Science, Universiti Brunei Darussalam. He did his Ph.D. in Information Security from the Royal Holloway College, London. His present research interests include computer security, cryptography, high performance computing, and machine learning. Further info on his homepage: https://expert.ubd.edu.bn/ghani.naim. He can be contacted at email: ghani.naim@ubd.edu.bn.